\pdfoutput=1

\documentclass[11pt]{article}

\usepackage{EACL2023}

\usepackage{times}
\usepackage{latexsym}

\usepackage{microtype}
\usepackage{array}
\usepackage{pifont}
\usepackage{tabularx}
\usepackage{adjustbox}
\usepackage{multirow}
\usepackage{enumitem}
\usepackage{xspace}
\usepackage{tcolorbox}
\usepackage{booktabs,amsfonts,dcolumn}
\usepackage{hyperref}
\usepackage{url}
\usepackage{amsmath,amsthm,amsfonts,amssymb,bm,stmaryrd,bbm}
\usepackage{makecell}
\usepackage{graphicx}
\usepackage{xcolor,colortbl}
\usepackage{color,soul}
\usepackage{float}

\usepackage[T1]{fontenc}

\usepackage[utf8]{inputenc}

\usepackage{microtype}

\usepackage{inconsolata}

\newcommand\tf[1]{\textbf{#1}}
\newcommand\ttt[1]{\texttt{#1}}

\newcommand{\mcorr}{m_{\text{corr}}}
\newcommand{\mpred}{m_{\text{pred}}}

\renewcommand{\paragraph}[1]{\vspace{0.2cm}\noindent\textbf{#1}}

\newcommand{\tableindent}{~~}

\definecolor{ccon}{HTML}{fee9d4}
\definecolor{cood}{HTML}{d8f0d3}
\definecolor{cid}{HTML}{dae8f5}

\definecolor{gred}{HTML}{cc0200}
\definecolor{ggreen}{HTML}{38761c}

\newcommand{\up}{\textcolor{ggreen}{$\uparrow$}}
\newcommand{\down}{\textcolor{gred}{$\downarrow$}}

\newcommand{\neuwhite}{\textcolor{white}{$\downarrow$}}
\newcommand{\recipe}{efficient pre-training recipe}
\newcommand{\academicbert}{24hBERT}
\newcommand{\eight}{\mbox{80-10-10}}
\newcommand{\corrx}{\tilde x}

\definecolor{c1}{cmyk}{0,0.6175,0.8848,0.1490}
\definecolor{c2}{cmyk}{0.1127,0.6690,0,0.4431}
\definecolor{c3}{cmyk}{0.3081,0,0.7209,0.3255}
\definecolor{c4}{cmyk}{0.6765,0.2017,0,0.0667}
\definecolor{c5}{cmyk}{0,0.8765,0.7099,0.3647}

\newtcbox{\hlprimarytab}{on line, rounded corners, box align=base, colback=c3!10,colframe=white,size=fbox,arc=3pt, before upper=\strut, top=-2pt, bottom=-4pt, left=-2pt, right=-2pt, boxrule=0pt}
\newtcbox{\hlsecondarytab}{on line, box align=base, colback=red!10,colframe=white,size=fbox,arc=3pt, before upper=\strut, top=-2pt, bottom=-4pt, left=-2pt, right=-2pt, boxrule=0pt}

\newcommand{\dashifted}{\raisebox{0.5\depth}{\tiny$\downarrow$}}
\newcommand{\uashifted}{\raisebox{0.5\depth}{\tiny$\uparrow$}}
\newcommand{\da}[1]{{\small\hlsecondarytab{\dashifted{#1}}}}
\newcommand{\ua}[1]{{\small\hlprimarytab{\uashifted{#1}}}}

\title{Should You Mask 15\% in Masked Language Modeling?}

\author{Alexander Wettig$^*$\quad Tianyu Gao$^*$\quad Zexuan Zhong\quad Danqi Chen  \\
Department of Computer Science, Princeton University\\
\ttt{\{awettig,tianyug,zzhong,danqic\}@cs.princeton.edu}\\
}

\begin{document}
\maketitle
\renewcommand{\thefootnote}{\fnsymbol{footnote}}
\footnotetext[1]{The first two authors contributed equally.}
\renewcommand{\thefootnote}{\arabic{footnote}}


\begin{abstract}


Masked language models (MLMs) conventionally mask 15\% of tokens
due to the belief that more masking would leave insufficient context to learn good representations; 
this masking rate has been widely used, regardless of model sizes or masking strategies.
In this work, we revisit this important choice of MLM pre-training.
We first establish that 15\% is not universally optimal, and larger models should adopt a higher masking rate.
Specifically, we find that masking 40\% outperforms 15\% for BERT-large size models on GLUE and SQuAD.
Interestingly, an extremely high masking rate of 80\% can still preserve
95\% fine-tuning performance and most of the accuracy in linguistic probing,
challenging the conventional wisdom about the role of the masking rate.
We then examine the interplay between masking rates and masking strategies
and find that uniform masking requires a higher masking rate
compared to sophisticated masking strategies such as span or PMI masking.
Finally, we argue that increasing the masking rate has two distinct effects:
it leads to more corruption, which makes the prediction task harder; 
it also enables more predictions, which benefits optimization. 
Using this framework, we revisit BERT's 80-10-10 corruption strategy.
Together, our results contribute to a better understanding of MLM pre-training.\footnote{Our code and pre-trained models are publicly available at \hyperlink{https://github.com/princeton-nlp/DinkyTrain}{https://github.com/princeton-nlp/DinkyTrain}.}

\end{abstract}

\section{Introduction}
\label{sec:intro}

Pre-trained language models have transformed the landscape of natural language processing~\cite[][\emph{inter alia}]{devlin2019bert,liu2019roberta,raffel2020exploring,brown2020language}.
They are trained on vast quantities of text data and acquire rich and versatile language representations.
Compared to autoregressive models, which always predict the next token in a sequence,
masked language models (MLMs) like BERT~\cite{devlin2019bert} predict a masked subset of input tokens based on the remaining context and
are more effective on downstream tasks
due to their bidirectional nature.



BERT chooses a 15\% masking rate, based on the reasoning that models cannot learn good representations when too much text is masked,
and the training is inefficient when too little  is masked.
Surprisingly, this important choice  has been under-explored since 15\% masking is used ubiquitously by BERT's successors~\cite{liu2019roberta,joshi2020spanbert,lan2020albert,he2021deberta,levine2021pmimasking,izsak2021train}, regardless of model sizes, masking strategies and optimization recipes.\footnote{
Some exceptions are discussed in \S\ref{sec:related}.
}

In this work, we aim to understand the impact of masking rates.
We hypothesize that the optimal masking rate is not universally 15\%, but should depend on other factors.
First, we consider the impact of model sizes and establish that indeed larger models should adopt higher masking rates (\S\ref{sec:model_size}).
Specifically, we find that under an \recipe{}~\cite{izsak2021train}, 40\% outperforms 15\%
for BERT-large size models when fine-tuning on GLUE and SQuAD.

Interestingly, we observe that large models can still learn good representations even for very high masking rates:
if we mask as much as 80\% of input tokens and pre-trained models have a perplexity of more than $1000$,
the learned representations can still preserve more than 95\% of fine-tuning performance on downstream tasks,
compared to the default 15\% masking (Table~\ref{tab:maskrate_example}),
and show considerable performance in linguistic probing (\S\ref{sec:high_masking}).
This challenges common intuitions about masking rates and what models learn in MLM pre-training.

\definecolor{maskcolor}{HTML}{6e9feb}
\newcommand\mmm[1]{{\color{maskcolor}{{\hl{#1}}}}}
\setulcolor{black}

%

\newcommand\mask[1]{\adjustbox{cframe=gray}{#1}}

\sethlcolor{maskcolor}

\begin{table*}[t!]
    \centering
    \resizebox{0.99\textwidth}{!}{
        \begin{tabular}{clrlll}
            \toprule
            \multicolumn{3}{c}{\tf{Pre-training}} & \multicolumn{3}{c}{\tf{Fine-tuning}}\\
             \cmidrule(r){1-3}   \cmidrule(r){4-6}
            $m$ & Example & PPL & MNLI & QNLI & SQuAD\footnote{placeholder} \\ 
            \midrule
            15\% &  We study high \mmm{mask} ing rates \mmm{for} pre-training language models . & 17.7 & 84.2 & 90.9 & 88.0\\
            40\% & We study high \mmm{mask} \mmm{ing} rates \mmm{for} pre-\mmm{training} \mmm{language} models . & 69.4 & \tf{84.5} \ua{0.3} & \tf{91.6} \ua{0.7} & \tf{89.8} \ua{1.8}\\
            80\% & We \mmm{study} high \mmm{mask} \mmm{ing} \mmm{rates} \mmm{for} \mmm{pre-training} \mmm{language} models \mmm{.}  & 1141.4 & 80.8 \da{3.4} & 87.9 \da{3.0} & 86.2 \da{1.8} \\
            \midrule
            \multicolumn{3}{c}{\textit{Random initialization}} & 61.5 \da{22.7} & 60.9 \da{30.0} & 10.8 \da{77.2} \\

           \bottomrule
        \end{tabular}
    }
    \caption{
        Masked examples, validation perplexity (calculated in the same way as  \citealp{devlin2019bert}) of different masking rates on the one billion word benchmark~\cite{chelba2013one},
        and downstream task development performance (SQuAD: F1; accuracy for others).
        All the pre-trained models have a BERT-large architecture and are trained with the \recipe{} (\S\ref{sec:setup}).
        Full results are provided in Table~\ref{tab:main_table_dev}.
    }
    \vspace{-10pt}
    \label{tab:maskrate_example}
\end{table*}

We then focus on the strategy of which tokens to mask as an additional factor to the optimal masking rate of MLMs (\S\ref{sec:spanpmi}).
We find that different masking rates should be used with different masking strategies, and
the default uniform masking benefits more from higher masking rates
than more sophisticated masking strategies such as span~\cite{joshi2020spanbert,raffel2020exploring} and PMI masking~\cite{levine2021pmimasking};
when all methods are considered at their optimal masking rate, uniform masking achieves competitive performance.

\footnotetext[3]{
    For our SQuAD v1.1 experiments, we continue training the models with 512-token sequences for 2,300 steps and report F1. See Appendix~\ref{sec:app:exp} for more details.
}

Finally, we propose to dissect the masking rate into two factors (\S\ref{sec:corruption_and_prediction}):
the \textit{corruption rate}---how much of the context is corrupted (masked)---and the \textit{prediction rate}---how much of the tokens the model predicts on.
In MLMs, both are set to the masking rate.
However, these two factors have opposing effects:
higher prediction rates generate more training signals and benefit the optimization,
while higher corruption rates make the prediction task more challenging by providing less context.
To study the two factors independently, we design ablation experiments to disentangle corruption and prediction rates.
Thus, we can verify that models benefit from higher prediction rates
and suffer from more corruption.
Using this framework, we also discuss BERT's practice of predicting on original or random tokens (the \eight{} rule), and we find that models usually perform worse under this corruption strategy (\S\ref{sec:eighttenten}).


Together, our results demonstrate the overlooked impact of the masking rate in MLM pre-training
and our analysis disentangles its opposing effects of corruption and prediction.
We conclude by discussing the relation to work in other models and modalities (\S\ref{sec:related})
and by highlighting several new avenues for efficient MLM in the future (\S\ref{sec:discussion}).

\section{Background}

\subsection{Masked Language Modeling}
\label{sec:background}

We focus on the widely popular masked language modeling \citep{devlin2019bert}, a form of denoising-autoencoding, where a model is trained to restore a corrupted input sequence. Specifically, masked language models make independent predictions on the subset of masked tokens:
\begin{equation}
    \label{eq:mlm}
    \resizebox{.89\hsize}{!}{%
    $
    L(\mathcal{C})=\mathop{\mathbb{E}}\limits_{x\in \mathcal{C}} \mathop{\mathbb{E}}\limits_{\substack{\mathcal{M}\subset x\\|\mathcal{M}|=m|x|}}\left[\sum\limits_{x_i\in \mathcal{M}} \log p(x_i|\corrx)\right],
    $
    }
\end{equation}
where one masks $m$ (masking rate, typically 15\%) percentage of tokens from the original sentence $x$ and predicts on the masked token set $\mathcal{M}$ given the corrupted context $\corrx$ (the masked version of $x$).


Different masking strategies have been proposed to sample $\mathcal{M}$:
\citet{devlin2019bert} randomly choose from the input tokens with a uniform distribution; \citet{joshi2020spanbert} sample contiguous spans of text; \citet{levine2021pmimasking} sample words and spans with high pointwise mutual information (PMI). These advanced sampling strategies are adopted to prevent models from exploiting shallow local cues from uniform masking. 

MLMs can encode bidirectional context while autoregressive language models can only ``look at the past'', and thus MLMs are shown to be more effective at learning contextualized representations for downstream use~\cite{devlin2019bert}.
On the other hand, MLMs suffer a significant computational cost because it only learns from 15\% of the tokens per sequence, whereas autoregressive LMs predict every token in a sequence.
In this work, we focus on MLMs and study the effects of different masking rates on downstream performance.

\subsection{Experiment Setup}
\label{sec:setup}

\begin{figure*}[t]
    \centering
    \includegraphics[width=0.99\textwidth]{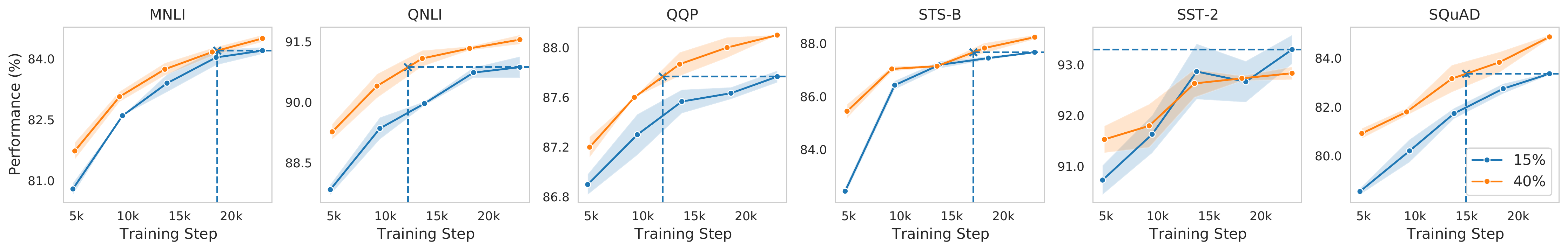}
    \caption{Downstream task development performance of \ttt{large} models trained with the \recipe{}, under masking rates of 15\% and 40\%.
    We highlight by the blue dotted line how long the 40\% model takes to achieve the same performance as the 15\% baseline;
    On QNLI and QQP, the 40\% model achieved the same performance with almost half the training time.
    }
    \vspace{-3pt}
    \label{fig:15_vs_40_reptask}
\end{figure*}

%
%
%

We build most of our experiments on a recent \recipe{}---the \academicbert{} recipe from \citet{izsak2021train}---by using which models can match BERT-base performance $6\times$ faster (tested on 8$\times$Titan-V).
This \recipe{} allows us to run a large amount of experiments in an academic setup.
\citet{izsak2021train} make the pre-training faster by using
a BERT-large architecture,
a larger learning rate (2e-3),
a larger batch size (4,096),
a shorter sequence length (128)\footnote{\citet{izsak2021train} only evaluate on GLUE tasks instead of SQuAD because of the short sequence length. We further train the model with 512 tokens for SQuAD in Table~\ref{tab:maskrate_example}.},
and fewer training steps.
We deviate from the \academicbert{} with a few simple changes:
\begin{enumerate}[leftmargin=0.03\textwidth]
    \item 
We adopt RoBERTa's BPE tokenizer~\citep{sennrich-etal-2016-neural,liu2019roberta}
rather than BERT's tokenizer
for it performs better in our preliminary experiments (see Appendix~\ref{sec:app:tokenizer}). 

    \item
Instead of adopting BERT's \eight{} token corruption strategy, we simply replace all the masked tokens with \ttt{[MASK]} by default.
We find that the \eight{} corruption strategy does not perform better for most downstream tasks, as discussed in
\S\ref{sec:eighttenten}.
\end{enumerate}

Following 24hBERT, we also do not perform next sentence prediction during pre-training, which was shown to hurt performance~\citep{liu2019roberta}.
We show  hyperparameters for the \recipe{} and a comparison to other recipes~\cite{devlin2019bert,liu2019roberta} in Appendix~\ref{sec:app:exp}.
For models of different sizes, masking rates, and masking strategies, we follow the same recipe as our preliminary experiments show that it still performs the best. 

We use fine-tuning downstream task performance as the measurement of how good the MLMs are,
since fine-tuning is the predominant way to use pre-trained MLMs in downstream use.
As evident from Table \ref{tab:maskrate_example},
pre-training metrics like perplexity do not correlate well with the downstream performance.
We describe our downstream fine-tuning setting and hyperparameters in Appendix~\ref{sec:app:exp}.

\begin{figure}[t]
    \centering
    \includegraphics[width=0.99\linewidth]{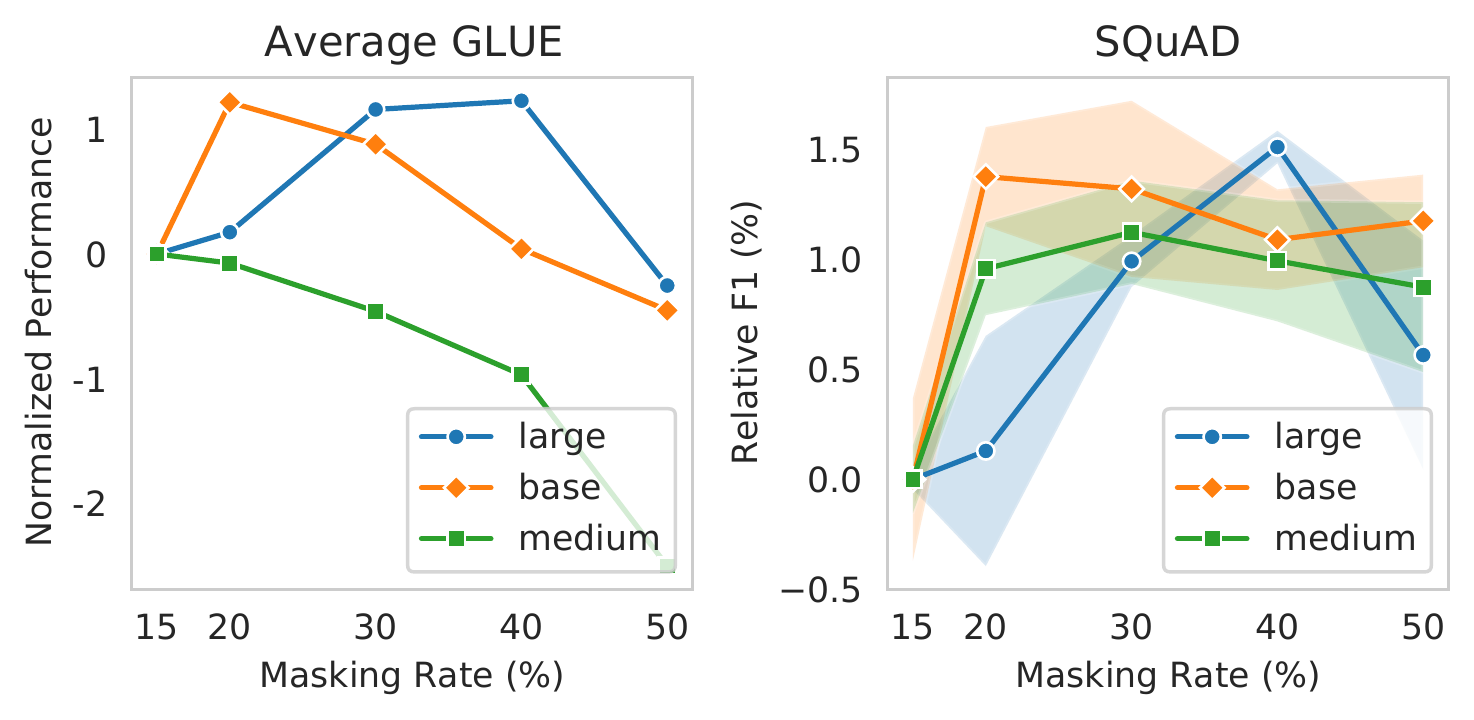}
    \caption{Impact of masking rates on different model sizes (\ttt{large}$>$\ttt{base}$>$\ttt{medium}).\footnote{placeholder} We see that larger models favor larger optimal masking rates.
    }
    \label{fig:model_size_mask_normalize_qqp}
    \vspace{-5pt}
\end{figure}
\footnotetext[5]{
    For each task and each model size, \emph{normalized performance} is calculated by $\frac{x-x_{15\%}}{\sigma}$ where $x_{15\%}$ is the performance of 15\% masking rate and $\sigma$ is the standard deviation across all masking rates.
    \emph{Relative F1} is the F1 score subtracted by the 15\% model F1.
}

\begin{figure*}[t]
    \centering
    \includegraphics[width=0.99\textwidth]{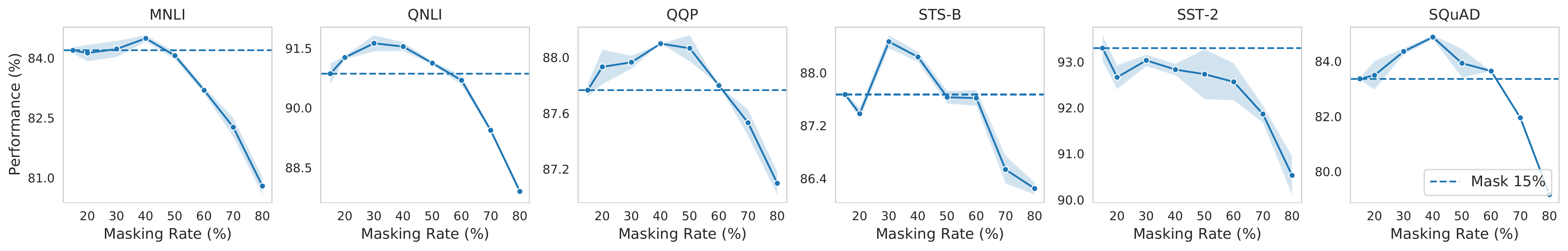}
    \caption{Impact of masking rates on \ttt{large} models with the \recipe{}.
    We see that on most tasks, higher masking rates outperform 15\%. 40\% is the optimal masking rate overall.
    }
    \label{fig:large_mask}
\end{figure*}

\section{Larger Models Can Benefit From Higher Masking Rates}
\label{sec:model_size}

\citet{devlin2019bert} choose the mysterious masking rate of 15\%,
for the belief that masking more leads to insufficient context to decode the tokens, and
masking fewer makes the training inefficient, and this masking rate has been viewed as a constant across different model sizes.
In this section, we train models of size \ttt{large} (354M parameters), \ttt{base} (124M parameters), and \ttt{medium} (51M parameters)
for masking rates varying from 15\% to 50\%. The model configurations are listed in Appendix~\ref{sec:app:models}.

\paragraph{Optimal masking depends on model sizes.}
\label{sec:model_size}
The impact of masking rate across the model sizes is summarized by Figure~\ref{fig:model_size_mask_normalize_qqp},
with detailed results given in Appendix~\ref{sec:app:models}.
We see that larger models possess higher optimal masking rates:
on average, under the \recipe{},
\ttt{large} models take 40\% as the optimal masking rate;
\ttt{base} models take  20\% and
\ttt{medium} models take 15\%.
This shows that larger MLM models favor higher masking rates.
We hypothesize that the additional capacity allows the large MLM to ``handle'' the more challenging task of
predicting many tokens given less context.




\paragraph{Large models learn faster with 40\% masking.}
We now compare the best performing masking rate 40\% to the conventional 15\% in more detail for our \ttt{large} model.
First, we plot how the downstream task performance changes with different training steps in Figure~\ref{fig:15_vs_40_reptask}.
For most tasks, we see that 40\% masking outperforms 15\% consistently during the course of training,
such that on QNLI and QQP, the 40\% model can achieve the same performance as the 15\% baseline with only half the training time.
We also report the test results in Table~\ref{tab:main_table_test},
where again masking 40\% outperforms 15\% with our \recipe{}.
However, the optimal masking rate can be task-dependent,
as SST-2 performs better with 15\% masking at the end of training.
We acknowledge that the optimal masking rate may also depend on the training recipe.
Since the \recipe{} uses a relatively small number of training steps,
we explore training for over $4\times$ more steps,
as well as training with a more expensive recipe from RoBERTa \citep{liu2019roberta},
and we find in Appendix~\ref{sec:app:long} that using a 40\% masking rate still performs well,
achieving similar performance to the 15\% masking rate. 
The experiments in the remaining sections of this paper are all based on large models.

\newcommand{\bsl}{\makebox[0pt][r]{\raisebox{0.05em}{$\bigstar\,$}}}

\begin{table*}[ht]
    \centering
    \resizebox{0.98\textwidth}{!}{
        \begin{tabular}{lcccccccccc}
            \toprule
             & \tf{MNLI-m/mm} & \tf{QNLI} & \tf{QQP} & \tf{RTE} & \tf{SST-2} & \tf{MRPC} & \tf{CoLA} & \tf{STS-B} &  \tf{SQuAD}\\
            \midrule
            Masking 15\%  & 84.2/83.4 &90.9 & 70.8 &73.5 & \tf{92.8} & 88.8 & \tf{51.8} & 87.3 & 88.0\\
            \bsl Masking 40\%  & \tf{84.7}/\tf{84.0} & \tf{91.3} & \tf{70.9} & \tf{75.5} &  92.6 & \tf{89.8} &  50.7 & \tf{87.6} & \tf{89.8} \\
            \bottomrule
        \end{tabular}
    }
    \caption{
    The test results on the GLUE benchmark with \ttt{large} models, the \recipe{} \citep{izsak2021train}, and with 15\% or 40\% masking rates.
    For RTE, MRPC, and STS-B we fine-tune  from the MNLI model following convention set by \citet{phang2018sentence}.
    For SQuAD v1.1, we take the same setting as Table~\ref{tab:maskrate_example}.
    }
    \label{tab:main_table_test}
    \vspace{-3pt}
\end{table*}

\section{MLMs in High-Masking Regimes}
\label{sec:high_masking}

The success of masking 40\% over 15\% motivates us to explore what happens at even larger masking rates.
Therefore, we pre-train additional \ttt{large} models with masking rates of up to 80\%.
We consider the question of what representations an MLM can learn with such limited input
as the last masked sentence in Table \ref{tab:maskrate_example}, which is hard to decipher even for a human.
While \citet{he2021masked} recently pioneered such high masking rates in the vision domain, and 
they reason that images are natural signals with heavy redundancy, while language is highly semantic and information-dense.
To our knowledge, nobody has examined such high masking rates in masked language modeling before.



\paragraph{MLMs learn with extreme masking.}
\label{sec:observation}
We first confirm in Table \ref{tab:maskrate_example} that the validation perplexity when pre-training with an 80\% masking rate is extremely high ($>$1,000),
which suggests that the MLM is unable to reconstruct corrupted inputs with independent token predictions.
Therefore our setting differs from vision, where good reproductions are possible with high masking rates \citep{he2021masked}.
Nevertheless, we find that MLMs can surprisingly still learn good representations:
Figure~\ref{fig:large_mask} shows the performance of the models fine-tuned on a range of tasks,
and we observe that pre-training with an 80\% masking rate can retain 95\% of fine-tuning performance,
which is substantially better than fine-tuning from a random initialization, which is reported in Appendix~\ref{sec:app:more}.

We hypothesize that MLMs at such high masking rates may be understood as a powerful skip-gram model~\citep{Mikolov2013EfficientEO},
e.g.,  masking 80\% of a 128 token sequence still learns skip-grams of length up to 26.
Furthermore, when compared to the simple word2vec model,
our Transformer models have access to positional information for each context token and prediction.




\begin{figure}[t]
    \centering
    \includegraphics[width=0.99\linewidth]{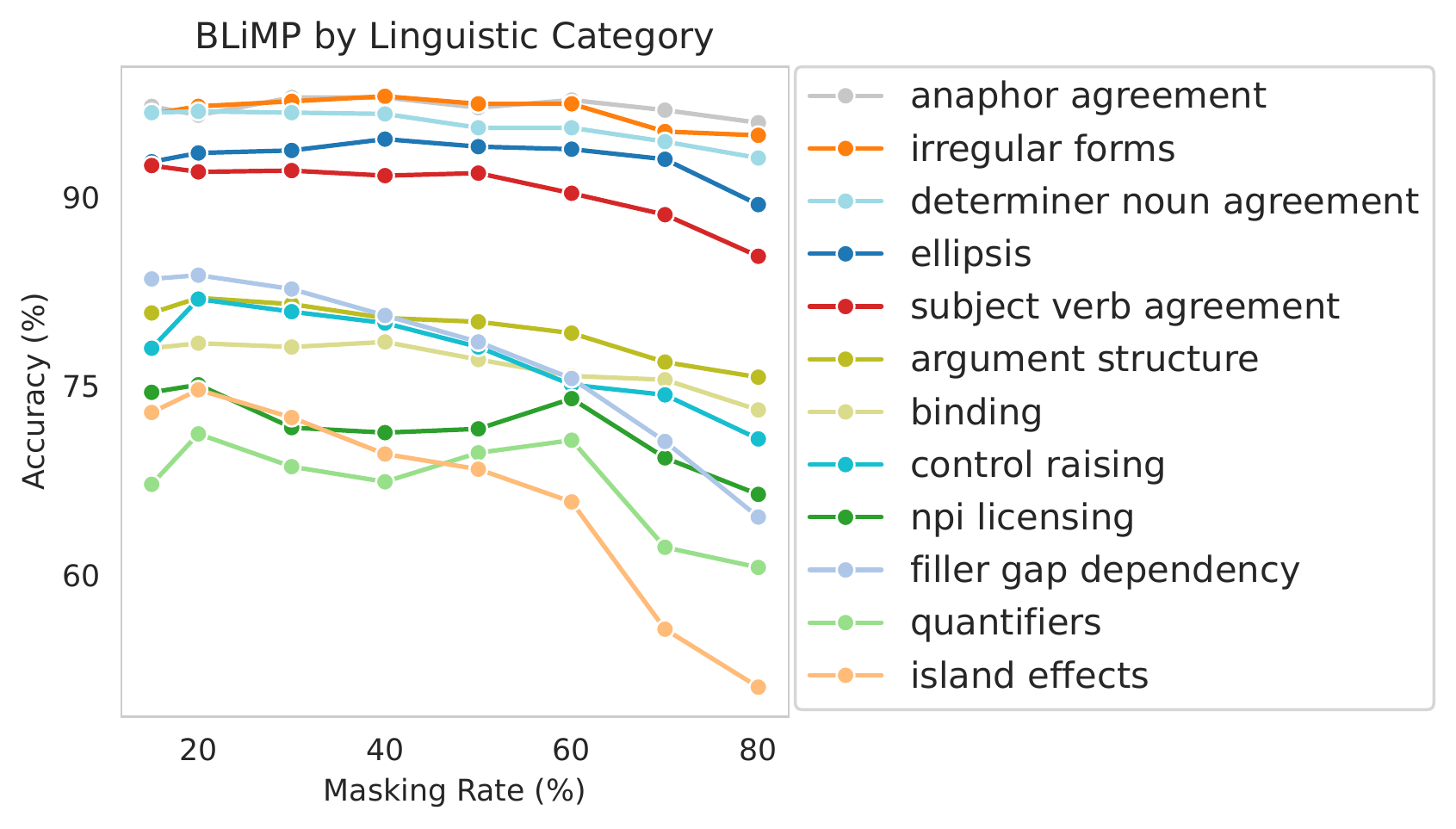}
    \caption{Evaluating our models on the BLiMP benchmark \citep{warstadt2019blimp} using pseudo log-likelihood scoring \citep{salazar-etal-2020-masked}.
    }
    \label{fig:blimp}
\end{figure}

\paragraph{Analysis of linguistic probing.}
Besides downstream performance, we study the models' linguistic abilities by evaluating them on the BLiMP benchmark \citep{warstadt2019blimp}.
We employ zero-shot pseudo log-likelihood scoring \citep{salazar-etal-2020-masked}, where a score is computed by masking each token individually,
which is a greater distributional shift from higher masking rates.
We show our results in Figure~\ref{fig:blimp}.
We find that most linguistic phenomena are acquired evenly across masking rates from 15\% to 60\%,
but they are still captured well by an MLM trained with 80\% masking---which on average preserves 90\% of the probing accuracy of the 15\% model baseline.
However, some categories such as filler gap dependencies and island effects
show clear trends that performance deteriorates with higher masking rates---although it remains unclear to what extent such linguistic knowledge is required by downstream tasks in GLUE \citep{sinha-etal-2021-masked}.
Overall, our results suggest that useful linguistic knowledge can be learned from a ``patchy'' training signal.



\section{Masking Rates vs. Masking Strategies}
\label{sec:spanpmi}

\citet{devlin2019bert,liu2019roberta} use uniform sampling for selecting which tokens to mask.
Subsequent work showed that adopting more sophisticated masking strategies---such as span masking or PMI masking---can outperform uniform masking on a range of downstream tasks~\cite{joshi2020spanbert,levine2021pmimasking}.
The argument for adopting advanced masking is that
uniform masking enables models to exploit shallow local cues \cite{levine2021pmimasking}.
An example is given by ``\ttt{[MASK]} Kong'': the model can easily predict ``Hong'' without using more context.
However, all the previous studies used a constant 15\% masking rate regardless of masking strategies, which raises the question of whether the conclusions still hold with a higher masking rate.



\begin{figure}[t]
    \centering
    \includegraphics[width=0.99\linewidth]{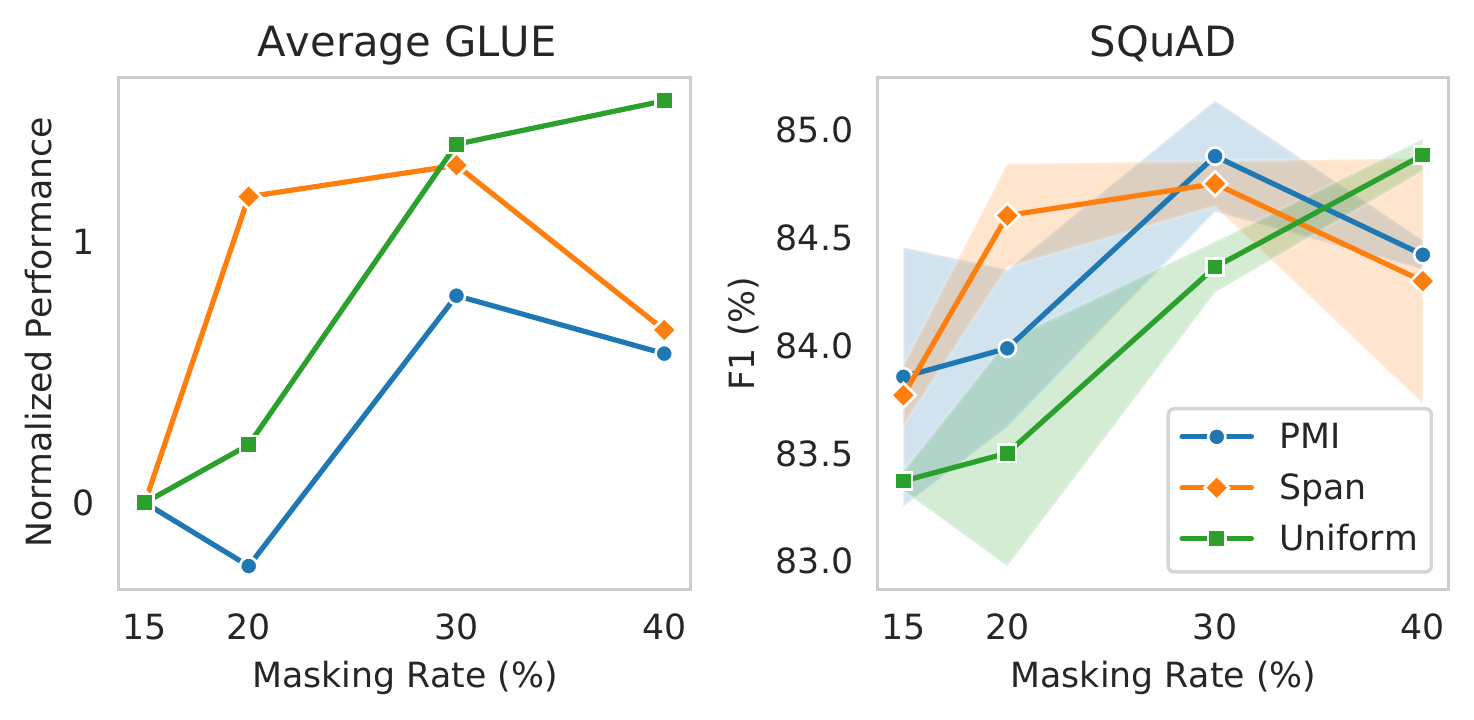}
    \caption{Performance of different masking strategies trained with different masking rates
    (\recipe{}, \ttt{large} models).
    }
    \label{fig:mask_strategy_mask_normalize_squad}
\end{figure}

We experiment with multiple masking strategies as an additional factor for the optimal masking rate in \ttt{large} models.  
Figure~\ref{fig:mask_strategy_mask_normalize_squad}
shows the results of
uniform masking,
T5-style span masking~\cite{raffel2020exploring}\footnote{Span maskings in \citet{raffel2020exploring} and \citet{joshi2020spanbert} differ in sampling procedures and we follow \citet{raffel2020exploring} for implementation simplicity.},
and PMI masking~\cite{levine2021pmimasking} under masking rates from 15\% to 40\%.
We see that
(1) for all masking strategies, the optimal masking rates are higher than 15\%;
(2) the optimal masking rates for span masking and PMI masking are lower than that of uniform masking; 
(3) when all strategies adopt the optimal masking rates,
the uniform masking achieves similar and even better results compared to the advanced strategies.
We also remark that, when masking with 15\%,
simply increasing the masking rate can be a more effective way to increase performance on SQuAD
than switching from uniform masking to another more advanced strategy.
More fine-grained results with these masking strategies are included in Appendix~\ref{sec:app:maskstrategy}.

\begin{figure}[t]
    \centering
    \includegraphics[width=0.99\linewidth]{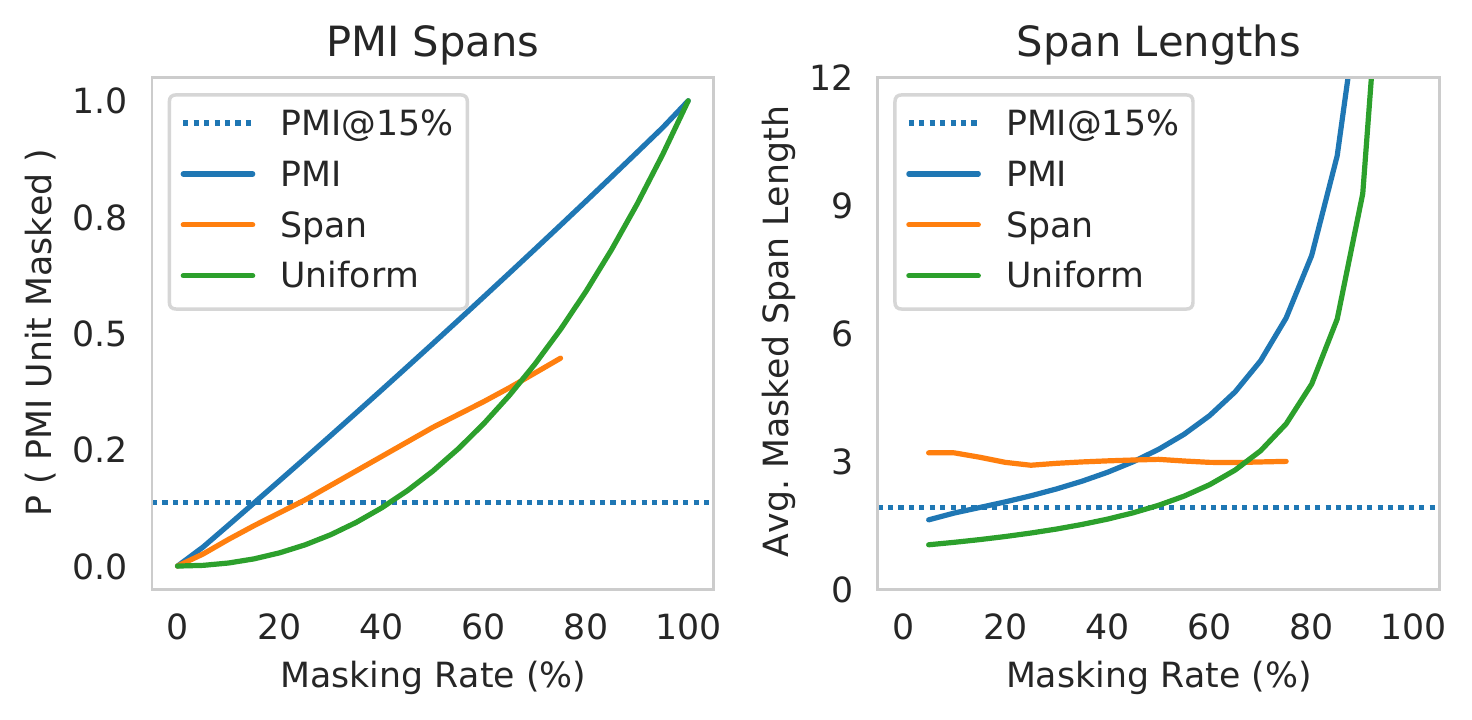}
    \caption{Higher masking rates increase the probability that an entire PMI span is masked (left) under different masking strategies. Uniform masking with a 40\% rate masks as many PMI spans as regular PMI masking at 15\%.
    Masks form longer spans for higher masking rates in uniform sampling, while the average length is fixed at $3$ for T5-style span masking (which cannot be enforced for very high masking rates).
    }
    \label{fig:masking_schemes}
\end{figure}

Interestingly, higher masking rates naturally increase the chance of masking neighbouring co-occuring tokens,
similar to the effect of the advanced masking strategies.
We consider the masked tokens over one epoch of training, and count the number of PMI n-grams (e.g., ``Hong Kong'') that were completely covered by different masking strategies.
Figure~\ref{fig:masking_schemes} shows that raising the masking rate from 15\% to 40\% results
in an 8-fold increase in the chance of masking a PMI n-gram under uniform masking and
gives a value comparable to PMI masking at 15\% masking rate.
Similarly, higher masking rates also make the masked tokens form longer spans.
However, at a given masking rate, uniform masking remains an easier task than span masking or PMI masking---it appears reasonable for uniform masking to admit a higher optimal masking rate for a given model capacity.





\section{Understanding Masking As Corruption and Prediction}
\label{sec:corruption_and_prediction}



In this section, we analyze
how masking rates affect the pre-training process of MLMs, through two distinct perspectives: task difficulty and optimization.
We identify that
the masking rate $m$ determines two import aspects of the pre-training problem:
the \emph{corruption} rate $\mcorr$
and
the \emph{prediction} rate $\mpred$.
$\mcorr$ is the proportion of tokens that are erased from the input sequence---typically by substituting \ttt{[MASK]}.
$\mpred$ is the proportion of tokens that the models predict, and each of those tokens contributes to the cross-entropy loss.


In Eq. (\ref{eq:mlm}), $\mcorr$ controls how much content is corrupted in $\corrx$ compared to the original sentence $x$, and $\mpred$ controls the number of predictions in the set $\mathcal{M}$.
Usually, both the corruption and the prediction rates are tied to the masking rate, i.e., $\mcorr=\mpred=m$, but they may impact representation quality differently.



\begin{table}[t]
    \centering
    \resizebox{1.02\linewidth}{!}{
        \begin{tabular}{cc|lllll}
            \toprule
            $\mcorr$ & $\mpred$ & \tf{MNLI} &  \tf{QNLI} & \tf{QQP} & \tf{STS-B} & \tf{SST-2} \\
            \midrule
            \cellcolor{cid}40\% & \cellcolor{cid}40\% & \cellcolor{cid}84.5$_{0.1}$ &	\cellcolor{cid}91.6$_{0.1}$  &	\cellcolor{cid}88.1$_{0.0}$ &	\cellcolor{cid}88.2$_{0.1}$   &	\cellcolor{cid}92.8$_{0.1}$  \\
            {\color{gray}40\%} & 20\% & 83.7\down &	90.6\down &	87.8\down &	87.5\down   &	92.9\up \\
            {20\%} & {20\%} & 84.1\down & 91.3\down & 87.9\down & 87.4\down & 92.7\down~ \\
            \midrule
            20\% & {\color{gray}40\%} & 85.7\up~  &	92.0\up~   &	87.9\down~ &	88.6\up~ &	93.4\up~ \\
            10\% & {\color{gray}40\%} & 86.3\up~  &	92.3\up~   &	88.3\up~  &	88.9\up~ &	93.2\up~ \\
            {\color{white}0}5\% & {\color{gray}40\%} & 86.9\up~ &92.2\up~ & 88.5\up~ & 88.6\up~  & 93.9\up~  \\
            \bottomrule
        \end{tabular}
    }
    \caption{Corruption vs. prediction. We take 40\% masking as the baseline model (standard deviation reported), disentangle $\mcorr$ and $\mpred$, and manipulate each independently. The trend is clear: more prediction helps and more corruption hurts. 
    }
    \label{tab:corruption_pred}
\end{table}

\paragraph{\boldmath{$\mcorr$} controls task difficulty.}
Masked language modeling attempts to learn a conditional probability distribution over the vocabulary given the corrupted context $p(\cdot \mid \corrx)$ during pre-training.
If a larger proportion of the input is corrupted, a token prediction is conditioned on fewer context tokens, making predictions harder and more uncertain.

\paragraph{\boldmath{$\mpred$} affects optimization.}
Predicting more means the model learns from more training signals,
so higher prediction rates boost the model performance.
From another perspective,
each prediction at each masked token leads to a loss gradient, which is averaged to optimize the weights of the model.
Averaging across more predictions has a similar effect to increasing the batch size, which is proved to be beneficial for pre-training~\cite{liu2019roberta}.


\paragraph{Experiments.}
In masked language modeling, both $\mcorr$ and $\mpred$ are determined by the overall masking rate.
To study how $\mcorr$ and $\mpred$ affect the downstream performance independently,
we design a simple ablation experiment to disentangle them:

\begin{enumerate}[wide, labelwidth=!, labelindent=0pt]
    \item If $\mpred < \mcorr$,
    we mask $\mcorr$ of tokens and
    only make predictions on $\mpred$ of the tokens.
    This can be  implemented without  additional cost.
    For example, with $\mcorr=40\%$ and $\mpred=20\%$, we mask 40\% and only predict on 20\% tokens.
    \item If $\mpred > \mcorr$,
    we duplicate each sequence $\lceil \frac{\mpred}{\mcorr} \rceil$ times and
    mask disjoint sets of $\mcorr$ of the tokens in different sequences.
    For example, with $\mcorr=20\%$ and $\mpred=40\%$, for each sentence, we do twice $20\%$ masking on different tokens and predict on all the masked tokens---this leads to a $20\%$ corruption but a $40\%$ prediction on each sequence.
    Note that this ablation takes  $\lceil \frac{\mpred}{\mcorr} \rceil$ times longer because we do multiple passes on every sequence, and is not efficient in practice.
\end{enumerate}

Table~\ref{tab:corruption_pred} shows the ablation results with disentangled $\mcorr$ and $\mpred$.
We see that
(1) fixing the $\mcorr$ as $40\%$, lowering the $\mpred$ from $40\%$ to $20\%$ results in a consistent drop on downstream tasks, showing that more predictions lead to better performance;
(2) fixing the $\mpred$ as $40\%$, lowering the $\mcorr$ leads to consistently better performance, suggesting that lower corruption rates make the pre-training task easier to learn and are better for pre-training.
Though we see that the performance gain by lowering $\mcorr$ from $10\%$ to $5\%$ is much smaller than that by lowering $\mcorr$ from $40\%$ to $20\%$, suggesting a diminishing marginal return of reducing the corruption rate.
(3) comparing $\mcorr=20\%,\mpred=20\%$ and $\mcorr=40\%,\mpred=40\%$, we see that the gain brought by more predictions transcends the drawback of more corruption, leading to better performance.


The ablation shows that when we tune the masking rate, we are tuning the corruption rate and the prediction rate together, which have antagonistic effects.
The final outcome is decided by which rate weighs more---the model benefits from higher masking rates if the hindrance brought by high corruption is surpassed by the advantage from predicting more.
Many factors may affect the balance between the two---for example, model sizes and masking strategies as we discussed in \S\ref{sec:model_size} and \S\ref{sec:spanpmi}.






\section{Revisiting BERT's Corruption Strategy}
\label{sec:eighttenten}



\begin{table}[t]
    \centering
    \resizebox{0.99\linewidth}{!}{
        \begin{tabular}{llllll}
            \toprule
            & \tf{MNLI} &  \tf{QNLI} &     \tf{QQP} & \tf{STS-B} & \tf{SST-2} \\
            \midrule
            \cellcolor{cid}40\% mask & \cellcolor{cid}84.5$_{0.1}$ &	\cellcolor{cid}91.6$_{0.1}$  &	\cellcolor{cid}88.1$_{0.0}$ &	\cellcolor{cid}88.2$_{0.1}$   &	\cellcolor{cid}92.8$_{0.1}$  \\
            \tableindent +5\% same     & 84.2\down~ &  91.0\down~ &	87.8\down~ &	88.0\down~ &	93.3\up~ \\
            \tableindent w/ 5\% rand  & 84.5\neuwhite~ & 91.3\down~ &87.9\down~ & 87.7\down~  & 92.6\down~ \\
            \tableindent w/ \eight & 84.3\down~ &  91.2\down~ &	87.9\down~ &	87.8\down~ &93.0\up~ \\
            \bottomrule
        \end{tabular}
    }
    \caption{
        Impact of substituting masks with random/same tokens. 
        ``+5\% same'': do extra 5\% same token predictions.
        ``w/ 5\% rand'': use mask for 35\% mask tokens and random tokens for 5\% .
        ``w/ \eight'': for the 40\% masked tokens, 10\% are same token predictions and 10\% are random token corruptions.
    }
    \label{tab:same_ran}
\end{table}

\citet{devlin2019bert} suggest that it is beneficial to replace 10\% of \ttt{[MASK]} tokens with the original token (\textit{same token predictions}) and 10\% with random tokens (\textit{random token corruptions}).
Since then, this \eight{} rule has been widely adopted in almost all the MLM pre-training work~\cite{liu2019roberta,joshi2020spanbert,he2021deberta}.
The motivation is that
masking tokens create a mismatch between pre-training and downstream fine-tuning, and
using original or random tokens as an alternative to \ttt{[MASK]} may mitigate the gap.
With our corruption and prediction framework,
we revisit the two kinds of mask replacements in the \eight{} rules and
empirically verify whether they are beneficial to downstream performance.





\paragraph{Same token predictions.}
The loss from same token predictions is very small and should be regarded as an auxiliary regularization.
Thus, same token predictions should
neither count towards the corruption nor to the prediction---they do not corrupt the input and contribute little to learning.




\paragraph{Random token corruptions.}
Replacing with random tokens contribute to corruption and prediction rate, as
the input is corrupted and the prediction task is non-trivial.
In fact, we find that the loss is slightly higher on random tokens compared to \ttt{[MASK]}, as (1) the model needs to decide for all tokens whether the information at the input is from a corruption or not, and (2) predictions need to be invariant to large changes in the input embeddings.

\paragraph{Ablation experiments.}
We adopt the $m=40\%$ model using only \ttt{[MASK]} replacements as the baseline, on top of which we add three models:
\begin{enumerate}[wide, labelwidth=!, labelindent=0pt]
    \item ``+5\% same'': we mask 40\% of tokens but predict on 45\% of tokens. 
    Adding same token predictions does not change $\mcorr$ or $\mpred$.
    \item ``w/ 5\% random'': we mask 35\% of tokens and randomly replace another 5\% of tokens, predicting on 40\% in total.
    \item ``\eight{}'': the original BERT recipe. 
    Due to same token predictions, $\mcorr=\mpred=36\%$.

\end{enumerate}


As shown in Table \ref{tab:same_ran},  we observe that same token predictions and random token corruptions
deteriorate performance on most downstream tasks.
The \eight{} rule performs worse than simply using all \ttt{[MASK]}---with the exception of SST-2, where same token predictions are beneficial.
Overall, our results suggest that in the fine-tuning paradigm,
the model can adapt to full, uncorrupted sentences, regardless of the use of alternative corruption strategies in pre-training.
Therefore, we suggest to use only \ttt{[MASK]} for MLM pre-training.
We also present an analysis based on information flow~\cite{voita-etal-2019-bottom} in Appendix \ref{sec:app:informationflow}.









\section{Related Work}
\label{sec:related}

\paragraph{Masking rates and masking strategies.}
There exist a few works on studying the impact of masking rates, among which
\citet{liao-etal-2020-probabilistically} show that dynamically sampling the masking rate from 0\% to 100\% for each sequence can improve MLM's downstream performance as well as the ability as a generation model.
On the other hand, masking strategies are heavily explored
for both pre-training \cite{joshi2020spanbert,raffel2020exploring,levine2021pmimasking}
and intermediate pre-training \cite{ye-etal-2021-influence} without considering the effect of masking rates.

\vspace{2pt}
\paragraph{``Unrealistic'' MLM training.}
A recent line of work shows that linguistically implausible MLM objectives can achieve competitive or non-trivial downstream performance, e.g., training with shuffled word order \citep{sinha-etal-2021-masked},
with randomly generated sequences \citep{krishna-etal-2021-pretraining-summarization}, or predicting only the first character of masked tokens \citep{yamaguchi-etal-2021-frustratingly, alajrami-aletras-2022-pre}.
These studies echo our findings that even an ``unrealistical'' high masking rate can still lead to good downstream results.


\vspace{2pt}
\paragraph{Masking in other language models.}
Besides MLMs, there are other pre-training schemes, 
namely
autoregressive language models~\cite{radford2018improving,brown2020language} and
sequence-to-sequence (seq2seq) language models~\cite{raffel2020exploring,lewis-etal-2020-bart}.
Similar to MLMs, seq2seq models corrupt text with a  masking rate,
but they predict  with an autoregressive decoder and are fine-tuned in different ways;
\citet{song2019mass} also point out that masking rates control whether seq2seq models are closer to encoder-only MLMs (masking less) or decoder-only autoregressive LMs (masking more).
Thus, we expect the masking rate studies in seq2seq models to draw a different conclusion from ours \cite{raffel2020exploring,tay2022unifying}.
Besides, \citet{tay2022scale} show that
pre-training metrics are not correlated with downstream performance, echoing our findings that perplexity does not correlate with fine-tuning results.


ELECTRA~\cite{clark2020electra} uses a smaller MLM to fill in 15\% of the blanks and trains a model to distinguish whether a token was generated by the MLM or not.
Despite the complicated training procedure,
the main motivation of ELECTRA is to improve the training efficiency by predicting on 100\% of tokens.
Interestingly, we find that the corruption rate in ELECTRA becomes very low towards the end of training---the average corruption rate is roughly only 7\%, but the replacements are ``hard'' negatives generated by the smaller MLM.
We leave the study of its connection to corruption and prediction rates as future work.


\vspace{2pt}
\paragraph{Masking in other modalities.}
Recently, a number of works extend MLM training to images and videos and demonstrate strong pre-training results~\cite{he2021masked,zhou2022image,feichtenhofer2022masked, tong2022videomae} .
They adopt extremely high masking rates (e.g., 75\% on images and 90\% on videos) compared to their language counterparts, with the argument that images and videos are highly information redundant.
\citet{baevski2020wave} propose a similar style masked model in speech and adopt a masking rate of around 50\%.




\vspace{3pt}
\section{Conclusion \& Discussion}
\label{sec:discussion}

In this work, we conduct a comprehensive study on the masking rates of MLMs. We discover that 15\% is not universally optimal, and larger models should adopt a higher masking rate. We also find that masking strategies should be considered together with masking rates, and uniform masking needs a higher masking rate than more sophisticated masking strategies. We gain a better understanding of masking rates by disentangling them as corruption rates and prediction rates and analyze the 80-10-10 corruption strategy that are widely used in BERT models. Based on our findings, we discuss the implications of high masking rates and future directions of efficient MLM pre-training:

\vspace{2pt}
\paragraph{Implications on higher masking rates.}
A direct takeaway from our findings is that larger models may adopt higher masking rates for better sample efficiency.
Figure~\ref{fig:15_vs_40_reptask} shows that a large model with 40\% masking can achieve comparable results to a 15\% baseline on several tasks with half the training time.
Larger models also exhibit faster convergence for a given computational budget:
\citet{li2020train} suggest it is more efficient to train larger models for fewer steps, as opposed to training smaller models for longer.
This can be combined with higher masking rates for better sample efficiency.


\vspace{2pt}
\paragraph{Separating masked and unmasked tokens.}
The training efficiency can potentially benefit from encoding masked and unmasked tokens separately, where masked tokens use a much lighter-weight module.
If a high masking rate is taken, this can significantly reduce the training cost due to the shorter input to the encoder. 
A similar approach has been explored by masked autoencoders in  vision~\cite{he2021masked}, where 75\% of the input patches are masked and removed from the  input of the heavy encoder to achieve a $4.1\times$ speedup. Recently, \citet{liao2022mask} have applied these architectural improvements to natural language pre-training, and together with a high masking rate can accelerate MLM by a third of the pre-training budget.
\paragraph{Disentangling corruption and prediction. }
Models perform better when trained with lower corruption rates and higher prediction rates.
However, in standard MLMs, those two factors are always tied to the masking rate.
Methods which can encode a sequence once
and then efficiently predict many small sets of masks, 
for example by manipulating the attention, 
could substantially accelerate masked language modeling pre-training.

\section*{Limitations}
(1)
Our analysis of masking rates applies to a specific type of pre-training method, masked language modeling.
We are also interested in studying masking rates in other pre-trained methods, e.g., seq2seq models and ELECTRA, and
leave it for future work.
(2) While we have shown how the optimal masking rate
depends on model size and masking strategy,
there may be additional factors, such as the vocabulary size, pre-training corpus or language family.
In particular, our experiments focus on English, but languages with different structural and morphological features may have lower or even higher optimal masking rates, or rely more on advanced masking strategies.
(3)
We consider a well-established yet relatively small set of downstream tasks, which do not benchmark domain-specific knowledge or more advanced reasoning skills.
(4) Due to the expensive nature of our pre-training experiments, we were not able to train multiple pre-trained models over multiple seeds.
(5) Finally, our findings point out  several promising directions but the paper primarily aims to study and understandthe impact of masking rates with respect to different factors.
We leave exploring better architectures and methods for efficient pre-training to future work.


\section*{Ethical Considerations}
Large language models can exhibit various kinds of stereotypes,
as they capture societal biases encoded in the training data.
These associations are not detected by standard GLUE or SQuAD evaluation.
We do not expect that simple modifications of masking rates can make progress towards solving these problems.
Language model pre-training is also computationally expensive, which comes at a significant environmental cost.
Furthermore, it makes re-production and follow-up research difficult within an academic context.
We reduce the computational requirements by following and promoting an efficient pre-training recipe and our findings point to future research for efficient MLM.


\section*{Acknowledgements}

We thank Sadhika Malladi and
the members of the Princeton NLP group for helpful discussion and valuable feedback.
Alexander Wettig is supported by a Graduate Fellowship at Princeton University.
This work is also supported by a Google Research Scholar Award.

\bibliography{custom}
\bibliographystyle{acl_natbib}

\clearpage
\appendix


\section{Experiment Setup}
\label{sec:app:exp}

\subsection{Pre-training}
\label{sec:app:exp_pretrain}

We implement our pre-training work based on \ttt{fairseq}~\cite{ott-etal-2019-fairseq}. To further speed up pre-training, we integrate the \ttt{DeepSpeed}~\cite{Rasley2020DeepSpeedSO} Transformer kernel for speedup.

We keep the other setting the same as the \academicbert{}~\cite{izsak2021train},
except that we use the RoBERTa tokenizer~\cite{liu2019roberta} and we do not adopt the \eight{} rule.
We train our model on the English Wikipedia and BookCorpus~\cite{zhu2015aligning}.
We want to emphasize that using pre-layernorm~\cite{shoeybi2019megatron}
is essential for the high learning rate in \citet{izsak2021train} to work.
The hyperparameters for the \recipe{} are shown in Table~\ref{tab:pre_hyper}.
We train with 8 Nvidia GTX 2080 GPUs and use gradient accumulation to achieve the large batch sizes.

\begin{table}[h]
    \centering
    \small
    \begin{tabular}{cc}
        \toprule
        \tf{Hyperparameter} & \tf{Efficient pre-training recipe}\\
        \midrule
        Peak learning rate & 2e-3 \\
        Warmup proportion & 6\% \\
        Batch size & 4,096\\
        Training steps & 23,000\\
        Sequence length & 128\\
        Architecture & \ttt{large} \\
        \bottomrule
    \end{tabular}
    \caption{Our pre-training hyperparameter settings. }
    \label{tab:pre_hyper}
\end{table}

\subsection{Downstream Task Evaluation}
\label{sec:app:exp_ft}

We fine-tune our model on the
GLUE benchmark~\cite{wang2019glue}, including SST-2~\cite{socher2013recursive_sst-2}, CoLA~\cite{warstadt2019neural_cola}, MNLI~\cite{williams2018broad_mnli}, QNLI~\cite{rajpurkar2016squad}, RTE~\cite{dagan2005pascal_rte1,bar2006second,giampiccolo2007third_rte3,bentivogli2009fifth_rte4}, MRPC~\cite{dolan2005automatically_mrpc}, QQP\footnote{\url{https://www.quora.com/q/quoradata/}} and STS-B~\cite{cer2017semeval_sts-b},
and the SQuAD v1.1~\cite{rajpurkar2016squad} dataset.
For each dataset we run three random seeds and average the results.
We apply grid search for the GLUE datasets, as shown in Table~\ref{tab:glue_hyper}.
For SQuAD,
we use a learning rate of 1e-4, a batch size of 16, and train for 2 epochs.
For both GLUE and SQuAD we use a linear scheduling for learning rates.

\begin{table}[h]
    \centering
    \small
    \resizebox{0.98\linewidth}{!}{
    \begin{tabular}{lc}
        \toprule
        \tf{Hyperparameter} & \tf{MNLI, QNLI, QQP}  \\
        \midrule
        Peak learning rate & \{5e-5, 8e-5\} \\
        Batch size &32  \\
        Max epochs & \{3, 5\} \\
        \midrule
        & \tf{RTE, SST-2, MRPC, CoLA, STS-B} \\
        \midrule
        Peak learning rate &  \{1e-5, 3e-5, 5e-5, 8e-5\} \\
        Batch size & \{16, 32\} \\
        Max epochs & \{3, 5, 10\}\\
        \bottomrule
    \end{tabular}}
    \caption{Grid search hyperparameters for GLUE tasks.}
    \label{tab:glue_hyper}
\end{table}

For all the results in the paper, we report accuracy for MNLI, QNLI, RTE, SST-2; we report F1 score for QQP, MRPC, and SQuAD; we report Matthew's correlation for CoLA and Spearman's correlation for STS-B.

For the SQuAD results in Table~\ref{tab:maskrate_example} and Table~\ref{tab:main_table_test},
we further train the models for 2300 steps (10\% of the training) with a sequence length of 512, a learning rate of 5e-4, and a warmup rate of 10\%.
For other tables and figures, we present the SQuAD results without further pre-training, and the absolute numbers are lower because of the short pre-training sequence length.
For some of the figures in the paper, we only show the results of MNLI, QNLI, QQP, STS-B, SST-2, and SQuAD due to limited space. Those tasks are selected because they have larger training set and the results are more reliable.
We always show the development results in all our figures and tables except Table~\ref{tab:main_table_test}, where we report the test numbers for GLUE tasks.

\section{Different Masking Rates: Full Results}
\label{sec:app:more}


\begin{table*}[t]
    \centering
    \resizebox{0.98\textwidth}{!}{
        \begin{tabular}{lcccccccccc}
            \toprule
             & \tf{MNLI-m/mm} & \tf{QNLI} & \tf{QQP} & \tf{RTE} & \tf{SST-2} & \tf{MRPC} & \tf{CoLA} & \tf{STS-B} &  \tf{SQuAD}\\
            \midrule
            Masking 15\% & 84.2/84.6 & 90.9 & 87.8 & \tf{67.3} & \tf{93.3} & \tf{77.0} & 59.2 & 87.7 & 88.0\\
            Masking 40\%  & \tf{84.5}/\tf{84.8} & \tf{91.6} & \tf{88.1} & 67.0 & 92.8 & 76.9 & \tf{61.0} & \tf{88.2} & \tf{89.8} \\
            Masking 80\% & 80.8/81.0 & 87.9 & 87.1 & 58.6 & 90.5 & 72.1 & 38.7 & 86.3 & 86.2 \\
            \midrule
            Random initialization$^\dagger$ & 61.5/61.2 & 60.9 & 70.7 & 49.6 & 80.0 & 45.4 & 11.9 & 17.5 & 10.8\\
            \bottomrule
        \end{tabular}
    }
    \caption{
    The development results on the GLUE benchmark with \ttt{large} models, the \recipe{}, and with 15\%, 40\%, or 80\% masking rates. The SQuAD development results are attained with the same continuous training as in Table~\ref{tab:maskrate_example}.
    Compared to the random initialization model, 80\% masking rates clearly learn good representations for downstream tasks, despite having a very high perplexity.
    $\dagger$: The random initialization models are trained with the same fine-tuning hyperparameters as pre-trained models, thus they could be undertrained.
    }
    \label{tab:main_table_dev}
\end{table*}

Table~\ref{tab:main_table_dev} shows the performance of 15\%, 40\% and 80\% masked models on all GLUE tasks and SQuAD. We can see that 80\% masking largely preserves the downstream performance and 40\% outperforms 15\% on most tasks.


\begin{table*}[h]
    \centering
    \resizebox{0.98\textwidth}{!}{
        \begin{tabular}{lccccccccc}
            \toprule
            & \tf{MNLI-m/mm} & \tf{QNLI} & \tf{QQP} & \tf{RTE} & \tf{SST-2} & \tf{MRPC} & \tf{CoLA} & \tf{STS-B} &  \tf{SQuAD}\\
            \midrule
            \multicolumn{10}{c}{Train longer with the \recipe{}}\\
            \midrule
            Masking 15\% & 87.47/87.02&	92.95&	88.40 & 69.93 & 94.07&82.50&61.00&  88.89&87.29\\
            Masking 40\% & 86.63/86.83&	93.13&	88.40 & 68.87 & 94.67&79.50&61.23&  89.60&87.16\\
            \midrule
            \multicolumn{10}{c}{Recipe from RoBERTa} \\
            \midrule
            Masking 15\% & 87.40/87.23&	93.04&	88.43 & 67.53 & 94.13&80.80&59.80&  90.05&90.72\\
            Masking 40\% & 87.30/87.03&	92.90&	88.83 & 67.63 & 94.10&63.90&56.07&  87.94&91.23\\
 
            \bottomrule
        \end{tabular}
    }
    \caption{Development  results of 15\%  vs 40\% masking with larger pre-training budget. We use the recipe from Table 3 in \citet{liu2019roberta}, and the \recipe{} with more training steps. See Table \ref{tab:long_hyper} for hyperparameters.
    }
    \label{tab:officialroberta}
\end{table*}

\begin{figure*}[h]
    \centering
    \includegraphics[width=\textwidth]{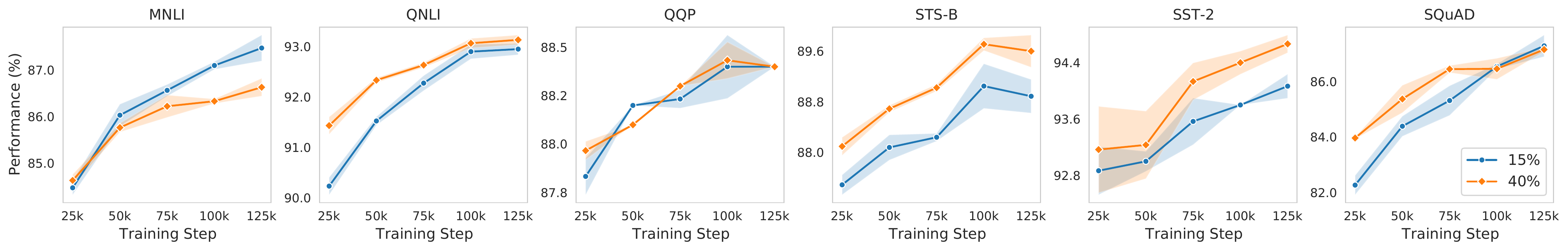}
    \caption{15\% vs 40\% masking rates with \ttt{large} models and the \recipe{} but trained longer.}
    \label{fig:train_long_15_vs_40}
\end{figure*}

\section{Tokenizer Comparison}
\label{sec:app:tokenizer}

Table~\ref{tab:tokenizer_comp} shows the performance of different tokenizers on downstream tasks.
We see that on most tasks RoBERTa tokenizer is better than BERT tokenizer.


\begin{table}[H]
    \centering
    \resizebox{0.99\linewidth}{!}{
        \begin{tabular}{lccccc}
            \toprule
            & \tf{MNLI-m/mm} & \tf{QNLI} &     \tf{QQP} & \tf{RTE}\\
            \midrule
            WordPieces & 84.3/\tf{84.9} & 90.8 & \tf{88.2} & 64.8\\
            BPE & \tf{84.5}/84.8 & \tf{91.6} & 88.1 &\tf{67.0} \\
            \midrule
            & \tf{SST-2} & \tf{MRPC} & \tf{CoLA} & \tf{STS-B} \\ 
            \midrule
            WordPieces & 92.5 &  75.5 & 56.6 &\tf{88.7}\\
            BPE& \tf{92.8} & \tf{76.9} & \tf{61.0} & 88.2\\
            \bottomrule
        \end{tabular}
    }
    \caption{
        Comparison between BERT's uncased WordPieces tokenizer and RoBERTa's BPE tokenizer. Both models are \ttt{large} and trained with the \recipe{} with a 40\% masking rate.
    }
    \label{tab:tokenizer_comp}
\end{table}

\section{Longer Training}
\label{sec:app:long}

\begin{table}[H]
    \centering
    \small
    \begin{tabular}{ccc}
        \toprule
        \tf{Hyperparameter} & \tf{Train longer} & \tf{RoBERTa}\\
        \midrule
        Peak learning rate & 2e-3 & 7e-4 \\
        Warmup proportion & 6\% & 6\%\\
        Batch size & 4,096 & 2,048\\
        Training steps & 125,000 & 125,000\\
        Sequence length & 128 & 512 \\
        \bottomrule
    \end{tabular}
    \caption{Comparison between our longer pre-training recipes and a recipe from RoBERTa~\cite{liu2019roberta}.}
    \label{tab:long_hyper}
\end{table}

To see that how the different masking rates perform with longer training, we modify the \recipe{} for longer steps.
We also experiment with a recipe used in the RoBERTa paper~\cite{liu2019roberta}.
Since the final RoBERTa models use more training data,
we refer to the recipe used in RoBERTa's ablation in its Table 3. 
Table~\ref{tab:long_hyper} shows the hyperparameters for the longer training, as well as a comparison to the RoBERTa's recipe. The major difference is that we train with much larger learning rate and only a sequence length of 128.

We train the models with 15\% and 40\% masking rates longer and evaluate them on downstream tasks.
Figure~\ref{fig:train_long_15_vs_40} shows the results.
We see that on most of the tasks, the trend that 40\% 
is better than 15\%
still holds, though the 40\% has a larger advantage when the training steps are limited.

We also train the model using a recipe from  RoBERTa  and present the results in Table~\ref{tab:officialroberta}.
We see that (1) on most tasks 40\% achieves comparable  results compared to 15\%; (2) our ``train longer'' results, which uses shorter sequences and larger learning rates, are comparable to the RoBERTa recipe results though with much shorter time.


%
%
%

\section{Results of Different Model Sizes and Masking Strategies}

\label{sec:app:models}

\begin{figure*}[h!]
    \centering
    \includegraphics[width=\textwidth]{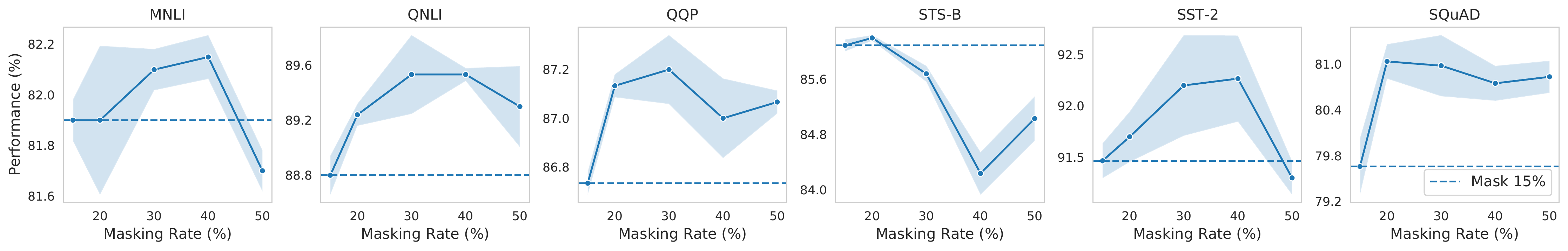}
    \caption{Results on selected downstream tasks with the \ttt{base} (124M parameter) model.}
    \label{fig:base}
\end{figure*}

\begin{figure*}[h!]
    \centering
    \includegraphics[width=\textwidth]{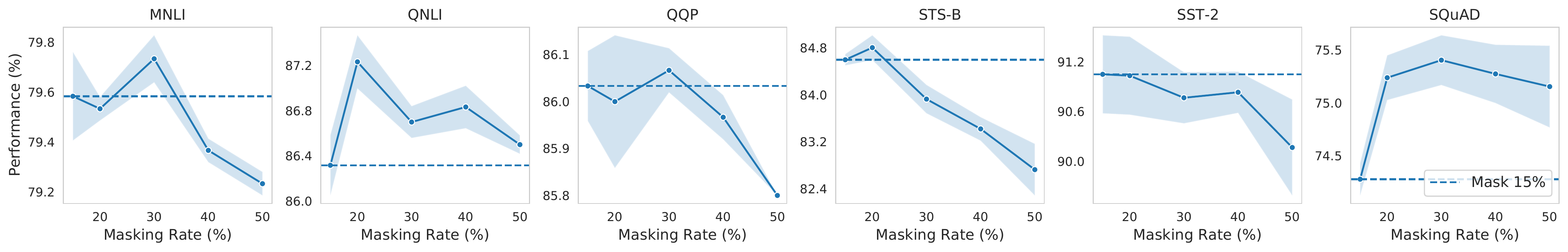}
    \caption{Results on selected downstream tasks with the \ttt{medium} (51M parameter) model.}
    \label{fig:medium}
\end{figure*}

We show the configurations of different model sizes in Table~\ref{tab:model_size}.
Figure~\ref{fig:base} and Figure~\ref{fig:medium} show the results of the \ttt{base} model and the \ttt{medium} model, which serve as complementary materials for Figure~\ref{fig:model_size_mask_normalize_qqp}.


\label{sec:app:maskstrategy}

\begin{figure*}[h!]
    \centering
    \includegraphics[width=\textwidth]{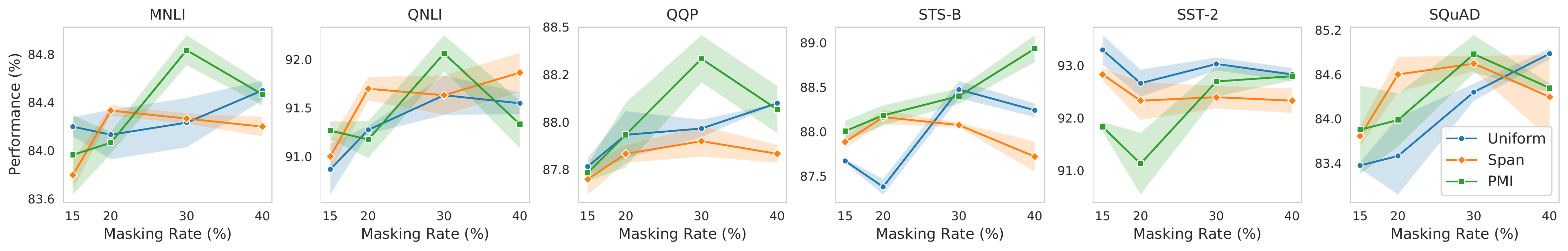}
    \caption{
        Comparison of different masking strategies on selected tasks. Models are trained with the \recipe{}, the \ttt{large} configuration, and several  masking rates.
    }
    \label{fig:seltask_mask_strategy}
\end{figure*}

Figure~\ref{fig:seltask_mask_strategy} shows the performance of
uniform masking,
T5-style span masking,
and PMI masking
on downstream tasks. This serves as a complementary material for Figure~\ref{fig:mask_strategy_mask_normalize_squad}.

\begin{table}[H]
    \centering
    \small
    \begin{tabular}{cccc}
        \toprule
         & \ttt{medium} & \ttt{base} & \ttt{large}\\
        \midrule
        \#Layers & 8 & 12 & 24\\
        \#Attention heads & 8 & 12 & 16\\
        Hidden size & 512 & 768 & 1024\\
        \bottomrule
    \end{tabular}
    \caption{Configurations of different model sizes.}
    \label{tab:model_size}
\end{table}

\section{Results on French MLM}
To validate our conclusions in a new setting, we conduct experiments on MLM on a corpus in French.
Similar to \citet{izsak2021train}, we pre-train on 2020 French Wikipedia and fine-tuned on French XNLI. We report accuracy averaged over 4 seeds, and make the observation that 40\% is better than 15\%.

\begin{table}[h]
    \centering
    \begin{tabular}{lcc}
    \toprule
     & \multicolumn{2}{c}{XNLI-fr} \\
     & valid     & test \\
    \midrule
    Masking 15\% & 78.3     & 77.3 \\
    Masking 40\%    & 78.9     & 77.5 \\
    \bottomrule
    \end{tabular}
    \caption{
We pre-train on 2020 French Wikipedia and fine-tuned on French XNLI. We report accuracy averaged over 4 seeds.
}
    \label{tab:my_label}
\end{table}

\begin{figure}[H]
    \centering
    \includegraphics[width=0.7\linewidth]{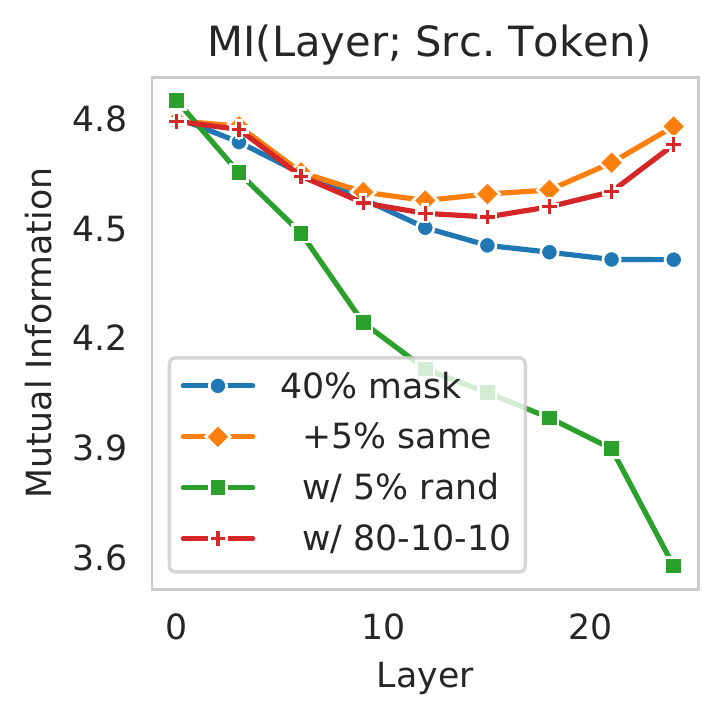}
    \vspace{-1mm}
    \caption{Mutual information between an input token and its intermediate representations for four different corruption strategies. See Table~\ref{tab:same_ran} for details on models.}
    \label{fig:mi_flow}
\end{figure}

\section{Information Flow Analysis}
\label{sec:app:informationflow}

To visualize the effect of these corruption strategies (the \eight{} rule), we follow \citet{voita-etal-2019-bottom}'s analysis of measuring mutual information between an input token and its intermediate representations.
Figure \ref{fig:mi_flow} shows that each model initially loses some information about the source token while acquiring information from the surrounding context.
Using same token predictions during pre-training leads to a ``reconstruction'' stage in the last few layers, as observed by \citet{voita-etal-2019-bottom}, whereby information about the source token is restored from the context.
However, this second stage is not present when same token predictions tokens are ablated:
the \texttt{[MASK]}-only baseline propagates contextual features only---and no reconstruction occurs.
This is more pronounced with random token corruption, where source information (that was less reliable during pre-training) is lost at a greater rate.
One consequence is that information about the input tokens can be more easily extracted when pre-training with same token predictions.
However, the reconstruction of the source tokens does not appear to be as important in the fine-tuning setting, as shown in our experiments in Table~\ref{tab:same_ran}.

\end{document}